\documentclass[letterpaper, 10 pt, conference]{ieeeconf}

\IEEEoverridecommandlockouts
\usepackage[english]{babel}
\usepackage{amsmath}
\usepackage{graphicx}
\usepackage[font=small]{caption}
\usepackage[colorlinks=true, allcolors=blue]{hyperref}
\usepackage{subcaption} 
\usepackage{algpseudocode}
\usepackage{algorithm}
\usepackage{multirow}
\usepackage{tabularx} 
\usepackage{comment}

\def\da{\downarrow}
\usepackage[dvipsnames,table]{xcolor}
\usepackage{pifont}
\definecolor{Gray}{gray}{0.9}
\definecolor{LightCyan}{rgb}{0.88,1,1}
\definecolor{LightPurple}{rgb}{0.83,0.83,0.93}
\definecolor{LightRed}{rgb}{1,0.83,0.83}
\definecolor{LightGreen}{rgb}{0.8,1.0,0.8}
\definecolor{LightOrange}{rgb}{1.0, 0.63, 0.48}
\definecolor{LightYellow}{rgb}{1.0, 1.0, 0.88}

\newcommand*\colourcheck[1]{%
  \expandafter\newcommand\csname #1check\endcsname{\textcolor{#1}{\ding{52}}}%
}
\colourcheck{blue}
\colourcheck{green}
\colourcheck{ForestGreen}

\usepackage{tikz}
\usetikzlibrary{external, positioning}
\usepackage{pgfplots}
\pgfplotsset{width=10cm,compat=1.17}

\title{Geometrically Plausible Object Pose Refinement using Differentiable Simulation}
\hypersetup{
  pdftitle={Geometrically Plausible Object Pose Refinement using Differentiable Simulation},
  pdfauthor={Anil Zeybek, Rhys Newbury, Snehal Dikhale, Nawid Jamali, Soshi Iba, Akansel Cosgun}
}

\begin{document}

\author{Anil Zeybek$^{3,4}$, Rhys Newbury$^{1,4}$, Snehal Dikhale$^{2}$, Nawid Jamali$^{2}$, Soshi Iba$^{2}$, Akansel Cosgun$^{1,4}$%
  \thanks{$^{1}$Monash University, Australia}%
  \thanks{$^{2}$Honda Research Institute, USA}%
  \thanks{$^{3}$Technical University of Munich, Germany}%
  \thanks{$^{4}$Physical AI, Australia}%
}

\maketitle
\thispagestyle{empty}
\pagestyle{empty}

\begin{abstract}
  State-of-the-art object pose estimation methods are prone to generating geometrically infeasible pose hypotheses. This problem is prevalent in dexterous manipulation, where estimated poses often intersect with the robotic hand or are not lying on a support surface. We propose a multi-modal pose refinement approach that combines differentiable physics simulation, differentiable rendering and visuo-tactile sensing to optimize object poses for both spatial accuracy and physical consistency. Simulated experiments show that our approach reduces the intersection volume error between the object and robotic hand by 73\% when the initial estimate is accurate and by over 87\% under high initial uncertainty, significantly outperforming standard ICP-based baselines. Furthermore, the improvement in geometric plausibility is accompanied by a concurrent reduction in translation and orientation errors. Achieving pose estimation that is grounded in physical reality while remaining faithful to multi-modal sensor inputs is a critical step toward robust in-hand manipulation.
\end{abstract}

\section{Introduction}


Robotic manipulation strategies are generally categorized as either model-based or model-free, depending on their reliance on a priori object-specific knowledge, such as CAD models \cite{kleeberger2020survey}. While model-free approaches offer superior generalization to novel objects \cite{newbury2023deep}, model-based methods are widely adopted in structured environments where object geometry is known. For rigid objects, these methods typically rely on a robust 6D pose estimation pipeline \cite{tremblay2018deep}.

Dexterous in-hand manipulation remains a formidable challenge. Current state-of-the-art systems often decouple the problem by utilizing model-based pose estimation for perception while focusing complexity on the actuation and control schemes \cite{chen2022system}. However, in-hand manipulation introduces severe occlusions caused by the robot's own geometry, which degrades visual pose estimation accuracy \cite{andrychowicz2020learning}. While recent work has integrated tactile sensing to mitigate these effects \cite{Dikhale2022}, many existing estimators still produce geometrically infeasible hypotheses. These include poses that lack sufficient support against gravity or exhibit physical interpenetration between the object and the robotic hand - intuitive constraints that are often ignored by pose estimators.


\begin{figure}[t!]
  \centering
  \includegraphics[trim=0.5cm 0.5cm 0.5cm 0.2cm, clip, width=1\linewidth]{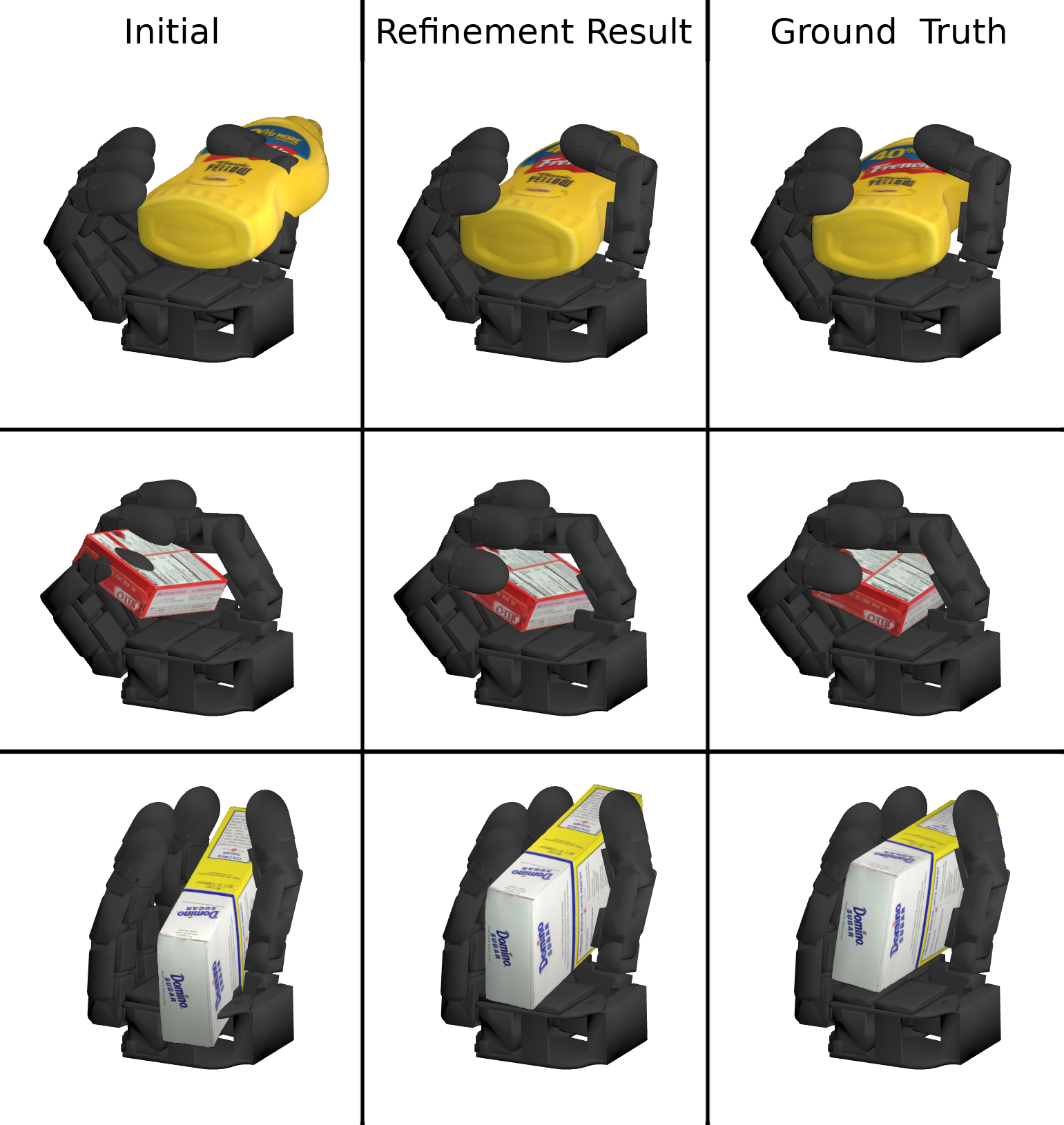}
  \caption{Given an initial estimate, our approach iteratively updates the object pose using differentiable simulation, the relative positioning of the estimated pose and the robotic hand as well as data from vision and tactile sensors.}
\end{figure}

Our approach leverages the known 3D geometry and joint configurations of the robotic platform to enforce physical consistency. By treating the object pose as a differentiable parameter, we utilize recent advances in differentiable physics simulation \cite{newbury2024Review} to iteratively refine pose hypotheses. Unlike traditional derivative-free optimization or trial-and-error refinement, differentiable simulation provides analytical gradients of the simulation output with respect to the 6D object pose. This allows for highly efficient gradient-based optimization within the search space.

Our formulation is influenced by \cite{handypriors}, who propose a pose estimation approach for human-object interaction, and \cite{turpin2022grasp}, who generate contact-rich grasps for high-DoF hands. Unlike these works, which optimize the hand configuration, we solve for the optimal 6D object pose relative to a fixed, known hand state. To ensure the refined pose remains grounded in reality, we introduce a multi-objective cost function that balances geometric plausibility (e.g., non-penetration and contact) with fidelity to visual and tactile sensor inputs. Our optimizer dynamically selects which gradient component to prioritize during each iteration based on the change in the loss functions. We evaluate our approach on a simulated dataset, demonstrating improvements in pose accuracy and physical feasibility compared to baseline.

\section{Related Work}

\subsection{Refinement Techniques for Pose Estimation}

6D object pose estimation is a cornerstone of computer vision, with extensive literature covering both RGB and RGB-D modalities. We refer readers to comprehensive surveys \cite{survey1, survey2} for a broad overview. Our work specifically focuses on post-estimation refinement, where an initial coarse hypothesis is iteratively improved.

\textbf{Geometric and Learning-based Refinement}: The Iterative Closest Point (ICP) algorithm \cite{ICP} remains a baseline for many systems \cite{xiang2017posecnn, Park2010, kehl2017ssd}. More recently, deep learning-based refinement has gained prominence \cite{wang2019densefusion, DPOD, BB8, Xu_2022_CVPR}. Large-scale render-and-compare methods such as FoundationPose \cite{wen2024foundationpose} and Co-op \cite{moon2025co} have demonstrated strong generalization, with the latter achieving state-of-the-art results across all seven BOP Challenge core datasets. While these methods achieve high accuracy, they operate purely on visual features and do not enforce any physical consistency constraints. Furthermore, they are designed for fully-observable object settings and cannot be directly applied to the severe occlusions and multi-modal sensor inputs characteristic of dexterous in-hand manipulation, making a direct comparison infeasible.

\textbf{Gradient-Based Optimization}: A powerful alternative is the `render-and-compare' paradigm \cite{diff_render1, diff_render2}. These methods optimize object poses by minimizing the pixel-wise or feature-wise difference between a rendered hypothesis and the observed scene via differentiable rendering. A close work to ours is ``HandyPriors'' \cite{handypriors}, which combines differentiable rendering with differentiable physics to ensure human-object interactions are both visually and physically consistent. We extend this by integrating tactile feedback, providing informative gradients in scenarios where visual data is occluded by the robotic hand.

\subsection{Physically Plausible Pose Estimation}

A scene is considered physically plausible if objects are neither ``floating'' in mid-air nor interpenetrating one another \cite{bauer2020physical}. Achieving this plausibility is a frequent objective in recent refinement literature \cite{bauer2022sporeagent, Bauer, mitash2018improving, kumar2021physically}.

Several strategies have been proposed to enforce these constraints. \cite{bauer2022sporeagent} utilized Reinforcement Learning to navigate the non-differentiable space of scene plausibility, while \cite{mitash2018improving} approached the problem through a physics-aware Monte Carlo Tree Search. DeepSimHO \cite{wang2023deepsimho} combines forward physics simulation with a learned gradient approximation to refine hand-object poses for stability. Most recently, PhysPose \cite{malenicky2025physpose} proposed a post-processing optimization that enforces non-penetration and gravitational support constraints on 6D pose estimates, achieving state-of-the-art results on the BOP benchmark.

Closest to our methodology is PhysPose \cite{malenicky2025physpose}, which also enforces physical constraints but treats refinement as a post-processing step without differentiable gradient flow. Unlike DeepSimHO \cite{wang2023deepsimho}, which learns to approximate simulator gradients, our approach uses fully differentiable simulation to obtain analytical gradients directly and unifies physical consistency with multi-modal sensor fidelity (visual and tactile) in a single optimization framework.

\section{Geometrically Plausible Pose Refinement}

We address the task of pose refinement by minimizing a multi-modal cost function that enforces both sensor fidelity and physical consistency. Our framework leverages recent advances in differentiable physics simulation to provide analytical gradients for an initial pose hypothesis $\mathbf{T} \in SE(3)$. The initial pose estimator we use is described in Sec.\ref{sec:initial_object_pose_estimation}. We derive four distinct gradient sources that are detailed in Sec.~\ref{sec:object_pose_refinement_gradients}. These are combined via a heuristic optimization strategy to iteratively refine the object pose (Sec.~\ref{sec:Heuristic_Optimization_Strategy}).

\begin{figure}[ht!]
  \centering
  \includegraphics[width=1\linewidth]{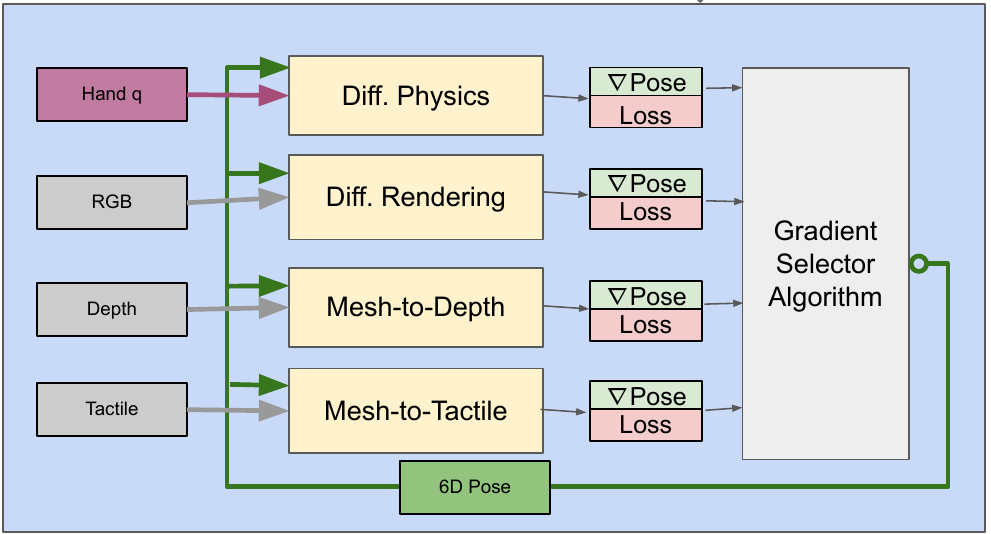}
  \caption{System Diagram. RGB-D and Tactile sensor inputs, and robot configuration is used to define 4 gradients. At each iteration, one of the gradients is selected according to a heuristic.}
  \label{fig: boxes}
\end{figure}

As depicted in Figure \ref{fig: boxes}, the sensor inputs are fed into each gradient function, and the loss of each method is calculated. We then use a heuristic algorithm to refine the pose. The rest of this section outlines these gradient methods and how we combine them to create our refinement approach.

\subsection{Initial Object Pose Estimation}
\label{sec:initial_object_pose_estimation}

While our refinement framework is agnostic to the source of the initial estimate, local gradient-based optimization requires a hypothesis within a reasonable convergence basin. We utilize a visuo-tactile estimation pipeline inspired by the architecture in \cite{Dikhale2022}, which processes fused RGB-D and tactile features through a specialized PointNet-based backbone to regress a 6D pose and a per-pixel confidence score. Their work modeled tactile sensors as proximity sensors, whereas we extract sparse point cloud data from active contact sensors, providing binary contact and force magnitude rather than continuous distance values. We also reduced the network size by fixing input image resolutions and pruning 1D convolutional layers, to facilitate rapid iteration and training.

\subsection{Multi-Modal Refinement Gradients}
\label{sec:object_pose_refinement_gradients}

\textbf{Differentiable Physics Simulation}: \cite{turpin2022grasp} proposed to use differentiable simulation to create grasp hypotheses, building on the Warp differentiable physics framework \cite{macklin2022warp}. They use a penalty-based formulation for contact forces and automatic differentiation for computing derivatives.  The contact forces are composed of a normal component proportional to the penetration depth and a frictional component computed using a Coulomb friction model. The penetration depth is calculated using a pre-computed Signed Distance Field (SDF). They compute a leaky gradient to work around inactive contacts and measure the grasp metric through simulation. While \cite{turpin2022grasp} optimize hand configurations, we fix the hand state and optimize only the object's 6D pose.


We observed that unconstrained physics optimization often over-corrects for translation while neglecting orientation. Since our initial estimator typically yields high-confidence positions but higher rotational error, we introduce an $L_2$ regularization term on the translational displacement $\Delta \mathbf{t}$:
\begin{equation}
  \mathcal{L}_{phys} = \mathcal{L}_{DP} + \lambda \|\Delta \mathbf{t}\|^2
\end{equation}
This forces the optimizer to prioritize rotational alignment to resolve interpenetrations and contact inconsistencies.



\textbf{Differentiable Rendering}: Similar to \cite{diff_render1,diff_render2}, we use differentiable rendering to optimize the pose of the object, using a `render-and-compare' approach. In our implementation, we utilize PyTorch3D \cite{ravi2020pytorch3d} as the rendering engine.

We consider the segmented RGB image from the camera and aim to align that with the simulated object pose. The simulated pose will not contain any occlusions; therefore, this approach will suffer greatly from occlusions caused by the robot's fingers. We use pixel-wise MSE over the entire RGB image as the loss function. By differentiating the loss function, we can propagate the gradients back to the pose of the object. This aims to create grasps which are similar to the ground truth in RGB space.

\textbf{Mesh-to-Depth Distance}: We convert the depth image into point cloud space, and calculate the distance between the estimated points and the object mesh at the current estimated pose. This is calculated using differentiable functions, therefore, we can use this function to optimize the pose of the object. The distance function is defined as:

\begin{equation}
  \mathcal{L}_{contact} = \frac{1}{|P|}\sum_{p}^{p \in P} \min_{t \in T} d(p, t)^2 + \frac{1}{|T|}\sum_{t}^{t \in T} \min_{p \in P} d(p, t)^2
  \label{pcd}
\end{equation}

where $P$ is the set of points in the point cloud, $T$ is the set of triangular faces in the object mesh, and $d$ represents a function that computes the distance between a triangle and a point. This aims to optimize the pose of the object with respect to the depth-image sensor input. Similar to differentiable rendering, this gradient is also degraded by occlusion.

\textbf{Mesh-to-Tactile Distance}: We again use Equation \ref{pcd}, however, we consider the contact points between the object and the fingers, rather than the depth image. We consider any tactile sensor with a force value larger than $\gamma$ to be in contact with the object. This aims to align the pose of the object with the tactile information. This will not be affected by the amount of occlusion, but it requires the object to be in contact with the hand to produce meaningful gradients.

\subsection{Heuristic Optimization Strategy}
\label{sec:Heuristic_Optimization_Strategy}

At each iteration, we dynamically weight each gradient source based on its historical effectiveness in minimizing the overall cost. Specifically, let $\Delta \mathcal{L}_i$ denote the change in the corresponding loss after executing a candidate step with gradient $\nabla_i$. The weight $\alpha_i$ assigned to gradient $i$ for the combined update is computed proportionally to $\max(0, -\Delta \mathcal{L}_i)$. Gradients that do not result in an improvement for their respective loss function are ignored ($\alpha_i = 0$). The final pose update at each step is a weighted sum of the contributing gradients, normalized so that $\sum_i \alpha_i = 1$. This process is repeated iteratively for a fixed number of steps or until no further improvements are seen in any of the loss functions. The goal is to prioritize the most effective gradients and dynamically adapt to the complex loss landscape.

Our experiments showed that the differentiable physics gradient can be much larger in magnitude and fluctuate, which can cause the optimizer to diverge or escape from a good local minimum. To prevent this, we implement a checkpointing mechanism that monitors the Differentiable Rendering loss ($\mathcal{L}_{DR}$). The state corresponding to the global minimum of $\mathcal{L}_{DR}$ is preserved as the final output, ensuring that physical refinement does not inadvertently ``drift'' the object away from the observed visual evidence.

\section{Experiments}

\subsection{Simulation Environment and Dataset}

We utilize the NVIDIA Isaac Sim platform \cite{isaac_sim1} to generate a high-fidelity synthetic dataset for in-hand manipulation. The simulated environment features a 6-DOF serial manipulator equipped with an Allegro four-fingered robotic hand. To simulate realistic tactile feedback, the hand is integrated with a virtual model of Xela tactile skin, providing 3-axis force measurements across 368 discrete sensors distributed throughout the palm and phalanges.

Following the methodology established by \cite{Dikhale2022}, we collect data by initiating random grasps on 9 objects from the YCB Dataset \cite{7251504}. For each object, the arm executes a randomized trajectory of 100 steps. We record 20,000 data points per object, each consisting of:

\begin{itemize}
  \item Tactile Feedback: 3-axis force vectors from 368 sensors ($0$ to $10,000$ internal units).
  \item Visual Data: Synchronized RGB and Depth images from a fixed camera.
  \item Proprioception: Full joint configurations for the arm and hand.
  \item Ground Truth: The 6D pose of the object relative to the camera frame.
\end{itemize}




\begin{figure}
  \centering
  \begin{subfigure}[t]{0.5\linewidth}
    \includegraphics[width=\linewidth]{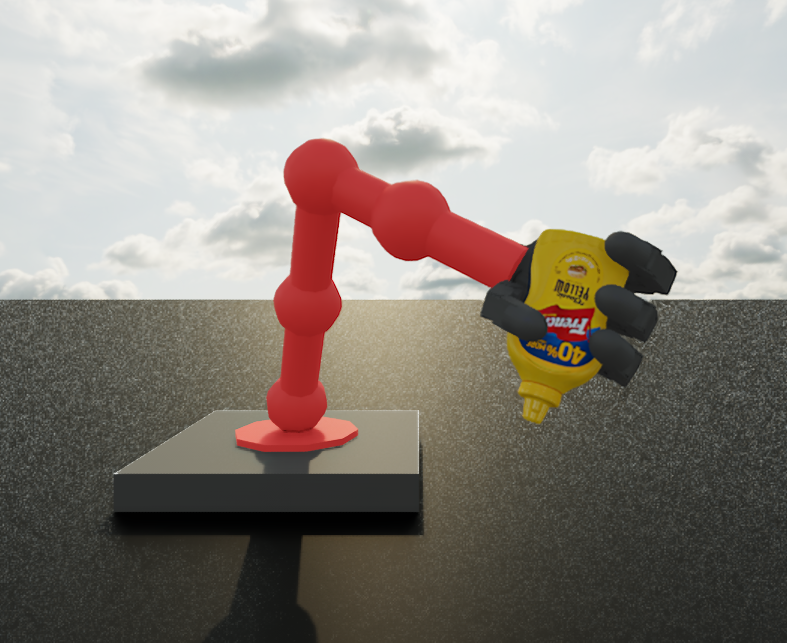}
    \caption{Serial manipulator grasping an object in simulation}
  \end{subfigure}
  \hfill
  \centering
  \begin{subfigure}[t]{0.4\linewidth}
    \includegraphics[width=\linewidth]{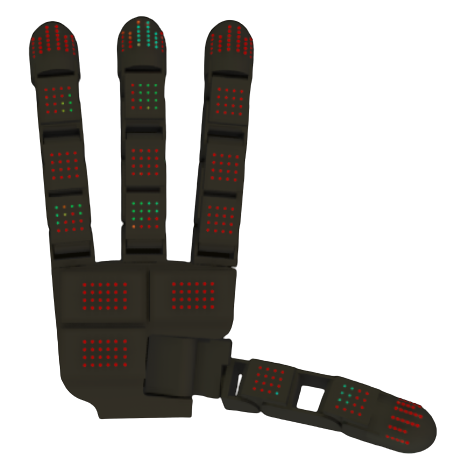}
    \caption{Xela robotic skin with 368 tactile sensors}

  \end{subfigure}
  \caption{We simulate a serial robotic arm, with an allegro robotic hand fitted with tactile sensors.}
  \label{fig:sim}
\end{figure}

\subsection{Training and Implementation Details}

The initial pose estimation network is trained independently for each of the 9 YCB objects using an 80/20 train-test split (16,000 training samples per model). Training is conducted for approximately 8 hours on a single NVIDIA RTX 3080 GPU. It is important to note that our refinement approach is inference-time optimization; it does not require additional training and operates directly on the output of the frozen estimation network.

\subsection{Evaluation Metrics}

We assess performance using two spatial accuracy metrics and two geometric consistency metrics. These align with the definition of physically plausible poses \cite{bauer2020physical}, where objects must neither ``float'' (measured by contact area) nor ``interpenetrate'' (measured by intersection volume).

\begin{figure}
  \centering
  \includegraphics[width=0.6\linewidth]{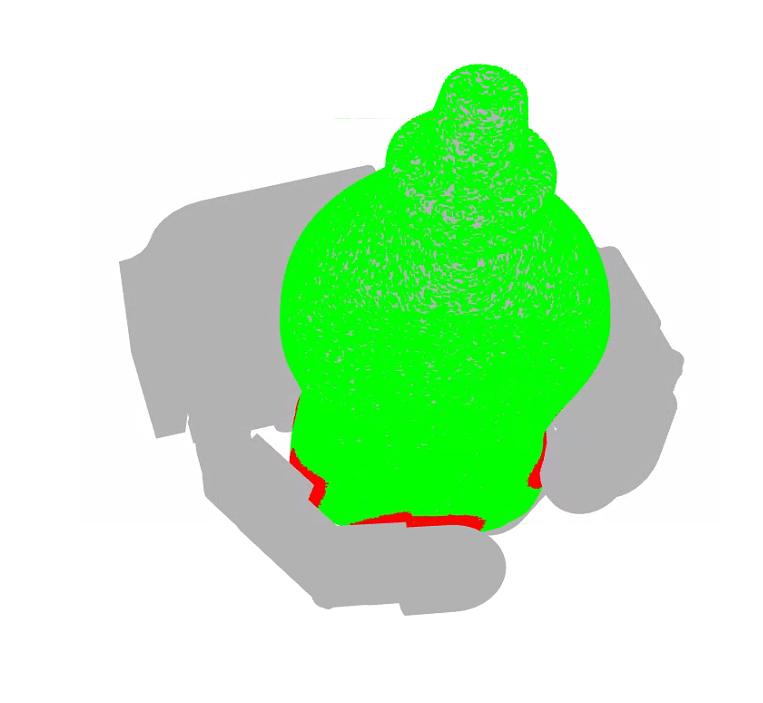}
  \caption{The Contact Area (CA) metric is shown in red between the fingers and the \textit{Mustard Bottle} object. 100k points are sampled on the surface of the object and contact is determined if the closest distance between the point and the robot is less than a threshold $\epsilon$.}
  \label{fig:ca}
\end{figure}

\begin{enumerate}
  \item Pose Accuracy:
    \begin{itemize}
      \item Position Error (PE): Calculated as the Euclidean distance between the refined translation vector and the ground truth.
      \item Orientation Error (OE): Following \cite{Dikhale2022}, we define the rotational error $\theta$ using the inner product of the estimated and ground truth quaternions:
        \begin{equation}
          \theta = \cos^{-1}(2\langle\hat{q},q\rangle^2 - 1)
        \end{equation}
    \end{itemize}

  \item Geometric Plausibility:
    \begin{itemize}
      \item Contact Area (CA): The Contact Area ($cm^2$) measures the surface-to-surface manifold between the object and the robotic hand. As illustrated in Figure \ref{fig:ca}, we compute this by uniformly sampling $10^5$ points on the object mesh and calculating the distance to the nearest point on the robot's surface. A point is considered "in contact" if the distance is within a threshold $\epsilon = 5\text{ mm}$. The total CA is the ratio of contact points to the total surface area.
      \item Intersection Volume (IV): To quantify physical interpenetration, we compute the overlapping volume ($cm^3$) between the hand and object meshes using the libigl library \cite{Panozzo2014LIBIGLAC}. This metric is a direct indicator of the physical impossibility of a pose hypothesis.
    \end{itemize}
  \item Deviation Metrics: To measure fidelity to the ground truth grasping state, we calculate the absolute difference between predicted and ground truth values for CA and IV ($|\Delta \text{CA}|$ and $|\Delta \text{IV}|$). While a high CA is often desired in grasp synthesis, our goal in pose estimation is to minimize the deviation from the actual observed contact manifold.
\end{enumerate}

\subsection{Baselines}

We evaluate our framework against a baseline of Iterative Closest Point (ICP), a standard post-processing step for point-cloud-based pose estimation \cite{xiang2017posecnn}. We also include an \textbf{ICP w/ Checkpointing} variant, which runs ICP for the full number of steps but retains the intermediate state that achieved the minimum differentiable rendering loss, preventing ICP from drifting too far from the visual observation. Each method is tested on 4,000 previously unseen data points per object. Our analysis includes:

\begin{enumerate}

  \item Ablation of individual gradients: Evaluating the contribution of Physics, Rendering, Depth, and Tactile signals.

  \item Iterative convergence: Assessing the effect of refinement steps ($N$) on final accuracy.

  \item Robustness to uncertainty: Evaluating performance when artificial noise is added to the initial pose.
\end{enumerate}

\section{Results}

Table \ref{table:default_results} shows the main results, with the initial object pose estimation approach explained in Sec.~\ref{sec:initial_object_pose_estimation}.

\setlength\tabcolsep{3.3 pt}
\begin{table}[!ht]
  \centering
  \begin{tabularx}{\linewidth}{|l|c|c|c|c|c|c|}
    \hline
    \rowcolor{LightCyan}
    \textbf{Method} & \textbf{PE}$\da$ & \textbf{OE}$\da$ & \textbf{CA} &  \textbf{IV} & \textbf{ $|\Delta \text{CA}|$}$\da$ &  \textbf{ $|\Delta \text{IV}|$}$\da$ \\ \hline
    \rowcolor{LightCyan}
    Scale/Unit & cm & deg & cm$^2$ & cm$^3$ & cm$^2$ & cm$^3$ \\ \hline
    \rowcolor{Gray}
    Ground Truth & - & - & 36.28 & 0.17 & 0 & 0 \\ \hline
    \rowcolor{LightRed}
    Initial Pose & 0.65 & 6.98 & 39.05 & 2.36 & \textbf{2.77} & 2.19\\ \hline
    \rowcolor{LightYellow}
    ICP & 3.46 & 43.01 & 65.26 & 28.73 & 28.98 & 28.56\\ \hline
    \rowcolor{LightYellow}
    ICP w/ Checkpointing & 1.30 & 13.77 & 48.31 & 8.71 & 12.03 & 8.54\\ \hline
    \rowcolor{LightGreen}
    Ours & \textbf{0.62} & \textbf{6.69} & 41.73 & 0.76 & 5.45 & \textbf{0.59} \\\hline
  \end{tabularx}
  \caption{Pose refinement results on 9 YCB objects. Best results are indicated in \textbf{bold}.}
  \label{table:default_results}
\end{table}

While initial estimates from the visuo-tactile backbone provide a strong baseline, our approach further reduces the mean position error by 4.6\% and orientation error by 4.1\%. Our approach yields a significant improvement in physical plausibility, achieving a 73.1\% reduction in intersection volume error compared to the initial pose estimate. The slight increase in $|\Delta\text{CA}|$ (from 2.77 to 5.45 cm$^2$, i.e., 7.6\% to 15.0\% of the ground truth CA of 36.28 cm$^2$) is a localized trade-off required to resolve deep interpenetrations, resulting in a contact manifold that is both numerically and physically more consistent with real-world grasping dynamics.

\subsection{Comparison to Individual Gradients}

In this experiment, we analyze whether the heuristic gradient selection approach is justified. We compare the complete approach to 4 alternatives, where each one executes only one of gradient functions. The results are shown in Table~\ref{table:individual_gradients}.

\setlength\tabcolsep{3.3 pt}
\begin{table}[!ht]
  \centering
  \begin{tabularx}{\linewidth}{|l|c|c|c|c|c|c|}
    \hline
    \rowcolor{LightCyan}
    \textbf{Method} & \textbf{PE}$\da$ & \textbf{OE}$\da$ & \textbf{CA} &  \textbf{IV} & \textbf{ $|\Delta \text{CA}|$}$\da$ &  \textbf{ $|\Delta \text{IV}|$}$\da$ \\ \hline
    \rowcolor{LightCyan}
    Scale/Unit & cm & deg & cm$^2$ & cm$^3$ & cm$^2$ & cm$^3$ \\ \hline
    \rowcolor{Gray}
    Ground Truth & - & - & 36.28 & 0.17 & 0 & 0 \\ \hline
    \rowcolor{LightRed}
    Initial Pose & 0.65 & 6.98 & 39.05 & 2.36 & \textbf{2.77} & 2.19\\ \hline
    \rowcolor{LightPurple}
    Mesh-to-Depth & 2.54 & 8.72 & 54.52 & 14.31 & 18.24 & 14.14\\ \hline
    \rowcolor{LightPurple}

    Mesh-to-Tactile & 3.24 & 9.51  & 90.63 & 32.68 & 54.35 & 32.51\\ \hline

    \rowcolor{LightPurple}
    Diff. Rendering & 0.85 & \textbf{6.65} & 43.99 & 3.60 & 7.71 & 3.43\\ \hline

    \rowcolor{LightPurple}
    Diff. Physics Sim & 1.26 & 6.59  & 39.01 & 1.26  & \textbf{2.73} & 1.09\\ \hline

    \rowcolor{LightPurple}
    Diff. Physics Sim w/CP & 0.78 & 7.62 & 43.06  & 0.62 & 6.78 & \textbf{0.45}\\ \hline

    \rowcolor{LightGreen}
    Ours & \textbf{0.62} & 6.69 & 41.73 & 0.76 & 5.45 & 0.59 \\\hline
  \end{tabularx}
  \caption{Comparison to individual gradients. Pose refinement results on 9 YCB objects. Best results are indicated in \textbf{bold}.}
  \label{table:individual_gradients}
\end{table}

Differentiable Physics with Checkpointing achieved the lowest $|\Delta \text{IV}|$; however, its position and orientation errors are worse than the initial estimate. This shows that physics alone ``pushes'' the object into a feasible spot, but ignores the sensor data. While Differentiable Rendering yields relatively low pose error, it results in high $|\Delta \text{IV}|$ and $|\Delta \text{CA}|$, representing a degradation in geometric plausibility. Both Mesh-to-Depth and Mesh-to-Tactile perform poorly when used in isolation. This is likely because these gradients are highly local and, without the global guidance of Rendering or Physics, they pull the object into massive interpenetrations to satisfy local point-distance costs.

Our combined method provides the best trade-off when all metrics are considered, achieving the lowest position error and a very competitive orientation error. It essentially uses the Physics gradient to ``fix'' the physical impossibilities of the visual methods without letting the object drift away from the camera's observation.

\subsection{Performance under High Uncertainty}

In this experiment, we injected artificial noise to the initial pose estimate. The goal of this experiment is to judge how our approach fares when the initial pose estimate is not very accurate. The results are shown in Table \ref{table:artificial_noise}.

\setlength\tabcolsep{3.3 pt}
\begin{table}[!ht]
  \centering
  \begin{tabularx}{\linewidth}{|l|c|c|c|c|c|c|}
    \hline
    \rowcolor{LightCyan}
    \textbf{Method} & \textbf{PE}$\da$ & \textbf{OE}$\da$ & \textbf{CA} &  \textbf{IV} & \textbf{ $|\Delta \text{CA}|$}$\da$ &  \textbf{ $|\Delta \text{IV}|$}$\da$ \\ \hline
    \rowcolor{LightCyan}
    Scale/Unit & cm & deg & cm$^2$ & cm$^3$ & cm$^2$ & cm$^3$ \\ \hline
    \rowcolor{Gray}
    Ground Truth & - & - & 36.28 & 0.17 & 0 & 0 \\ \hline
    \rowcolor{LightRed}
    Initial Pose & 2.21 & 15.25 & 61.34 & 21.57 & 25.06 & 21.40\\ \hline

    \rowcolor{LightYellow}
    ICP & 3.65 & 42.55 & 61.08 & 26.10 & 24.80 & 25.93 \\ \hline

    \rowcolor{LightYellow}

    ICP w/ Checkpointing  & 2.53 & 23.29  & 55.87 & 19.57 & 19.59 & 19.40 \\ \hline

    \rowcolor{LightGreen}
    Ours & \textbf{1.45} & \textbf{13.14} & 42.65 & 2.8 & \textbf{6.37} & \textbf{2.63}  \\\hline
  \end{tabularx}
  \caption{Refinement results when noise is added to the initial pose estimate (9 YCB objects). Best results are indicated in \textbf{bold}.}
  \label{table:artificial_noise}
\end{table}

The experimental results under artificial noise highlight a key strength of our proposed framework: Robustness to Poor Initialization. In the previous "no-noise" experiment, our method provided incremental gains in pose accuracy (4.6\% improvement in PE). However, when significant noise is introduced (almost tripling the baseline PE and doubling the baseline OE), the strength of our gradient-based refinement becomes far more evident. Our approach reduces the Position and Orientation Error by 34.4\% and 13.8\%, respectively. In contrast, standard ICP variants either diverge or fail to significantly improve the noisy hypothesis, often resulting in orientations worse than the initial estimate.

When geometric plausibility metrics are considered, our differentiable physics gradient effectively ``pushes'' the object out of the hand's geometry. We achieve a 87.7\% reduction in $|\Delta \text{IV}|$ compared to the initial pose. The fact that $|\Delta \text{CA}|$ remains low ($6.37\text{ cm}^2$ deviation vs $25.06\text{ cm}^2$ for the baseline) further suggests that our method successfully converges to a stable grasping state rather than a random non-colliding pose. This demonstrates that even when visual data is noisy or misleading, the physical consistency loss acts as a powerful regularizer that guides the object toward a plausible workspace.

\subsection{Effect of the Number of Refinement Steps}

We study the effect of the number of iterations, the effect of regularization and checkpointing. As the number of iterations increased, we found that the IV was consistently decreasing. Results are shown in Table~\ref{table:refinement_steps}.

\setlength\tabcolsep{3 pt}

\begin{table}[!ht]
  \center
  \begin{tabularx}{\linewidth}{|X|cccc|cccc|}
    \hline
    \rowcolor{LightCyan}
    & \multicolumn{4}{c|}{No Noise} & \multicolumn{4}{c|}{Artificial Noise} \\ \hline
    \rowcolor{LightCyan}
    \textbf{\# Iterations} & \textbf{PE}$\da$ & \textbf{OE}$\da$ & \textbf{CA}&  \textbf{IV}$\da$ & \textbf{PE}$\da$ & \textbf{OE}$\da$ & \textbf{CA} &  \textbf{IV}$\da$ \\ \hline
    \rowcolor{LightCyan}
    Scale/Unit & cm & deg & cm$^2$ & cm$^3$ & cm & deg & cm$^2$ & cm$^3$ \\ \hline
    \rowcolor{LightRed}
    Initial Pose & \textbf{0.65} & 6.44 & 47.10 & 2.00 & 2.31 & 15.15 & 73.06 & 27.55 \\ \hline
    \rowcolor{LightGreen}
    Ours (100 steps) & 0.80 & \textbf{5.88} & 40.65 & 0.12 &  1.75 & 13.74 & 47.85 & 1.80 \\ \hline
    \rowcolor{LightGreen}
    Ours (200 steps) & 0.92 & 5.97 & 40.21 & 0.08 & 1.73 & 13.17 & 44.23 & 0.95 \\ \hline
    \rowcolor{LightGreen}
    Ours (400 steps) & 1.06 & 6.50 & 40.30 & \textbf{0.06} & 1.73 & 12.42 & 42.13 & 0.54 \\ \hline
    \rowcolor{LightGreen}
    Ours (600 steps)  &  1.13 & 7.08 & \textbf{40.01} & \textbf{0.06} &  \textbf{1.72 }& 12.02 & 40.88 & 0.36 \\ \hline
    \rowcolor{LightGreen}
    Ours (1000 steps) & 1.32 & 8.41 & 40.48 & \textbf{0.06} &  1.74 & \textbf{11.85} & \textbf{40.33} & \textbf{0.22} \\\hline
  \end{tabularx}
  \caption{Analysis on the number of refinement steps. Only the Mustard object was used. Regularization and checkpointing was not active. Best results are indicated in \textbf{bold}.}
  \label{table:refinement_steps}
\end{table}

In the No Noise scenario, extended iterations ($N > 100$) lead to a monotonic decrease in Intersection Volume (IV), successfully resolving interpenetrations. However, without regularization, the pose begins to ``drift'' from the visual ground truth, as evidenced by the increase in PE. This behavior underscores the necessity of our Checkpointing system, which preserves the visually optimal pose while the physics engine continues to refine contact feasibility. Conversely, in the Artificial Noise scenario, the refinement process acts as a robust recovery mechanism. Increasing $N$ significantly reduces the orientation error and yields a significant reduction in IV (from 27.55 to 0.22 cm$^3$). For practical deployment, we select 400 iterations as our standard. This configuration provides a reasonable trade-off between accuracy and latency, requiring 60.6 seconds to converge compared to 154 seconds for 1000 iterations. While differentiable rendering remains the primary computational bottleneck, we anticipate that advancements in optimized rendering kernels will further reduce this latency.

\section{Discussion}

Our experiments demonstrate that multi-modal gradient-based refinement is a powerful tool for bridging the gap between perception and physical reality in robotic manipulation. The core challenge in in-hand pose estimation is not merely numerical accuracy, but geometric feasibility. Standard ICP-based methods and even high-capacity deep-learning estimators frequently generate implausible poses that physically interpenetrate the robotic geometry. By integrating a differentiable physics simulator, our framework reduces intersection volume ($IV$) errors by 73\% when the initial pose is relatively accurate and by over 87\% under conditions of high initial pose uncertainty. While our approach sometimes leads to a higher Contact Area than the ground truth, this metric shows significant improvement in noisy environments, suggesting a convergence toward more stable and realistic contact manifolds.

The simple yet effective gradient selection approach serves as a critical supervisor. As shown in our ablation study, pure differentiable physics can sometimes prioritize interpenetration resolution at the cost of visual fidelity, effectively ``pushing'' the object away from the true pose to satisfy a non-collision constraint. Our heuristic effectively identifies these divergent cases by monitoring the cost-landscape of the differentiable renderer, ensuring that physical ``sanitization'' does not come at the expense of sensor grounding. This synergy is confirmed by our results, where the refinement approach consistently decreased pose errors relative to both the initial estimate and the ICP baseline.

Our ablation study further confirms that geometric plausibility alone is not a sufficient objective; a pose can be physically valid but not matching the sensor reality. Our solution combines visual, depth, tactile, and physical gradients, ensuring that the final pose remains grounded in the sensor data while respecting the hard constraints of the physical world. The substantial improvement in the $CA/IV$ metric, a benchmark in grasp synthesis, also underscores the functional utility of our refined poses for downstream manipulation tasks.

Our evaluation is conducted entirely in simulation, and transferring the approach to real-world settings remains an important direction for future work. We note that the refinement pipeline is designed around inputs available on real robotic platforms --- RGB-D images, tactile contact readings, and known robot kinematics --- and the four gradient sources operate on standard sensor modalities and geometric representations (meshes, signed distance fields) without relying on simulator-specific state or privileged information. Nonetheless, real-world deployment will require addressing practical challenges such as sensor noise, calibration errors, and computational latency. Investigating these factors and validating sim-to-real transfer is a key priority for future research.

\section{Conclusion}

In this paper, we presented a novel framework for the 6D pose refinement of objects during dexterous in-hand manipulation. By leveraging differentiable physics simulation alongside differentiable rendering and visuo-tactile sensing, we successfully optimized object poses for both spatial accuracy and physical plausibility. Our approach significantly outperforms standard ICP baselines and provides a robust recovery mechanism for inaccurate initial pose estimates. Future work can focus on investigating the transferability of our method to real-world scenarios. Another potential avenue is improving the computational efficiency of the differentiable rendering pipeline to enable real-time, closed-loop pose refinement. For instance, replacing traditional mesh-based differentiable rendering with recent advances in 3D Gaussian Splatting for pose estimation \cite{jin20256dope,liu2025gscpr} offers a highly promising avenue for rapid, high-fidelity optimization.

\bibliographystyle{IEEEtran}
\bibliography{refs}

\begin{thebibliography}{10}
\providecommand{\url}[1]{#1}
\csname url@samestyle\endcsname
\providecommand{\newblock}{\relax}
\providecommand{\bibinfo}[2]{#2}
\providecommand{\BIBentrySTDinterwordspacing}{\spaceskip=0pt\relax}
\providecommand{\BIBentryALTinterwordstretchfactor}{4}
\providecommand{\BIBentryALTinterwordspacing}{\spaceskip=\fontdimen2\font plus
\BIBentryALTinterwordstretchfactor\fontdimen3\font minus \fontdimen4\font\relax}
\providecommand{\BIBforeignlanguage}[2]{{%
\expandafter\ifx\csname l@#1\endcsname\relax
\typeout{** WARNING: IEEEtran.bst: No hyphenation pattern has been}%
\typeout{** loaded for the language `#1'. Using the pattern for}%
\typeout{** the default language instead.}%
\else
\language=\csname l@#1\endcsname
\fi
#2}}
\providecommand{\BIBdecl}{\relax}
\BIBdecl

\bibitem{kleeberger2020survey}
K.~Kleeberger, R.~Bormann, W.~Kraus, and M.~F. Huber, ``A survey on learning-based robotic grasping,'' \emph{Current Robotics Reports}, 2020.

\bibitem{newbury2023deep}
R.~Newbury, M.~Gu, L.~Chumbley, A.~Mousavian, C.~Eppner, J.~Leitner, J.~Bohg, A.~Morales, T.~Asfour, D.~Kragic \emph{et~al.}, ``Deep learning approaches to grasp synthesis: A review,'' \emph{IEEE Transactions on Robotics}, 2023.

\bibitem{tremblay2018deep}
J.~Tremblay, T.~To, B.~Sundaralingam, Y.~Xiang, D.~Fox, and S.~Birchfield, ``Deep object pose estimation for semantic robotic grasping of household objects,'' \emph{arXiv preprint arXiv:1809.10790}, 2018.

\bibitem{chen2022system}
T.~Chen, J.~Xu, and P.~Agrawal, ``A system for general in-hand object re-orientation,'' in \emph{Conference on Robot Learning}.\hskip 1em plus 0.5em minus 0.4em\relax PMLR, 2022, pp. 297--307.

\bibitem{andrychowicz2020learning}
O.~M. Andrychowicz, B.~Baker, M.~Chociej, R.~Jozefowicz, B.~McGrew, J.~Pachocki, A.~Petron, M.~Plappert, G.~Powell, A.~Ray \emph{et~al.}, ``Learning dexterous in-hand manipulation,'' \emph{The International Journal of Robotics Research}, vol.~39, no.~1, pp. 3--20, 2020.

\bibitem{Dikhale2022}
S.~Dikhale, K.~Patel, D.~Dhingra, I.~Naramura, A.~Hayashi, S.~Iba, and N.~Jamali, ``Visuotactile 6d pose estimation of an in-hand object using vision and tactile sensor data,'' \emph{IEEE Robotics and Automation Letters (RA-L)}, 2022.

\bibitem{newbury2024Review}
R.~Newbury, J.~Collins, K.~He, J.~Pan, I.~Posner, D.~Howard, and A.~Cosgun, ``A review of differentiable simulators,'' \emph{IEEE Access}, 2024.

\bibitem{handypriors}
S.~Zhang, Y.-L. Qiao, G.~Zhu, E.~Heiden, D.~Turpin, J.~Liu, M.~Lin, M.~Macklin, and A.~Garg, ``Handypriors: Physically consistent perception of hand-object interactions with differentiable priors,'' in \emph{IEEE International Conference on Robotics and Automation (ICRA)}, 2024.

\bibitem{turpin2022grasp}
D.~Turpin, L.~Wang, E.~Heiden, Y.-C. Chen, M.~Macklin, S.~Tsogkas, S.~Dickinson, and A.~Garg, ``Grasp'd: Differentiable contact-rich grasp synthesis for multi-fingered hands,'' in \emph{European Conference on Computer Vision (ECCV)}, 2022.

\bibitem{survey1}
G.~Marullo, L.~Tanzi, P.~Piazzolla, and E.~Vezzetti, ``6d object position estimation from 2d images: a literature review,'' \emph{Multimedia Tools and Applications}, 2022.

\bibitem{survey2}
J.~Chen, L.~Zhang, Y.~Liu, and C.~Xu, ``Survey on 6d pose estimation of rigid object,'' in \emph{Chinese Control Conference}, 2020.

\bibitem{ICP}
P.~Besl and N.~D. McKay, ``A method for registration of 3-d shapes,'' \emph{IEEE Transactions on Pattern Analysis and Machine Intelligence}, vol.~14, no.~2, pp. 239--256, 1992.

\bibitem{xiang2017posecnn}
Y.~Xiang, T.~Schmidt, V.~Narayanan, and D.~Fox, ``Posecnn: A convolutional neural network for 6d object pose estimation in cluttered scenes,'' \emph{arXiv preprint arXiv:1711.00199}, 2017.

\bibitem{Park2010}
I.~K. Park, M.~Germann, M.~D. Breitenstein, and H.~Pfister, ``Fast and automatic object pose estimation for range images on the gpu,'' \emph{Machine Vision and Applications}, 2010.

\bibitem{kehl2017ssd}
W.~Kehl, F.~Manhardt, F.~Tombari, S.~Ilic, and N.~Navab, ``Ssd-6d: Making rgb-based 3d detection and 6d pose estimation great again,'' in \emph{IEEE international conference on computer vision}, 2017.

\bibitem{wang2019densefusion}
C.~Wang, D.~Xu, Y.~Zhu, R.~Mart{\'\i}n-Mart{\'\i}n, C.~Lu, L.~Fei-Fei, and S.~Savarese, ``Densefusion: 6d object pose estimation by iterative dense fusion,'' in \emph{IEEE/CVF conference on computer vision and pattern recognition}, 2019.

\bibitem{DPOD}
S.~Zakharov, I.~Shugurov, and S.~Ilic, ``Dpod: 6d pose object detector and refiner,'' in \emph{IEEE/CVF International Conference on Computer Vision (ICCV)}, 2019.

\bibitem{BB8}
M.~Rad and V.~Lepetit, ``Bb8: A scalable, accurate, robust to partial occlusion method for predicting the 3d poses of challenging objects without using depth,'' in \emph{IEEE International Conference on Computer Vision (ICCV)}, 2017.

\bibitem{Xu_2022_CVPR}
Y.~Xu, K.-Y. Lin, G.~Zhang, X.~Wang, and H.~Li, ``Rnnpose: Recurrent 6-dof object pose refinement with robust correspondence field estimation and pose optimization,'' in \emph{IEEE/CVF Conference on Computer Vision and Pattern Recognition (CVPR)}, 2022.

\bibitem{wen2024foundationpose}
B.~Wen, W.~Yang, J.~Kautz, and S.~Birchfield, ``Foundationpose: Unified 6d pose estimation and tracking of novel objects,'' in \emph{Proceedings of the IEEE/CVF conference on computer vision and pattern recognition}, 2024, pp. 17\,868--17\,879.

\bibitem{moon2025co}
S.~Moon, H.~Son, D.~Hur, and S.~Kim, ``Co-op: Correspondence-based novel object pose estimation,'' in \emph{Proceedings of the Computer Vision and Pattern Recognition Conference}, 2025, pp. 11\,622--11\,632.

\bibitem{diff_render1}
A.~Simpsi, M.~Roggerini, M.~Cannici, and M.~Matteucci, ``6 dof pose regression via differentiable rendering,'' in \emph{International Conference on Image Analysis and Processing}.\hskip 1em plus 0.5em minus 0.4em\relax Springer, 2022.

\bibitem{diff_render2}
I.~Shugurov, I.~Pavlov, S.~Zakharov, and S.~Ilic, ``Multi-view object pose refinement with differentiable renderer,'' \emph{IEEE Robotics and Automation Letters}, vol.~6, no.~2, pp. 2579--2586, 2021.

\bibitem{bauer2020physical}
D.~Bauer, T.~Patten, and M.~Vincze, ``Physical plausibility of 6d pose estimates in scenes of static rigid objects,'' in \emph{European Conference on Computer Vision}.\hskip 1em plus 0.5em minus 0.4em\relax Springer, 2020.

\bibitem{bauer2022sporeagent}
------, ``Sporeagent: reinforced scene-level plausibility for object pose refinement,'' in \emph{IEEE/CVF Winter Conference on Applications of Computer Vision}, 2022.

\bibitem{Bauer}
------, ``Verefine: Integrating object pose verification with physics-guided iterative refinement,'' \emph{IEEE Robotics and Automation Letters}, 2020.

\bibitem{mitash2018improving}
C.~Mitash, A.~Boularias, and K.~E. Bekris, ``Improving 6d pose estimation of objects in clutter via physics-aware monte carlo tree search,'' in \emph{IEEE International Conference on Robotics and Automation (ICRA)}, 2018.

\bibitem{kumar2021physically}
A.~Kumar, A.~R. Vaidya, and A.~G. Huth, ``Physically plausible pose refinement using fully differentiable forces,'' \emph{arXiv preprint arXiv:2105.08196}, 2021.

\bibitem{wang2023deepsimho}
R.~Wang, W.~Mao, and H.~Li, ``Deepsimho: Stable pose estimation for hand-object interaction via physics simulation,'' \emph{Advances in Neural Information Processing Systems}, vol.~36, pp. 79\,685--79\,697, 2023.

\bibitem{malenicky2025physpose}
M.~Malenick{\`y}, M.~C{\'\i}fka, M.~Fourmy, L.~Montaut, J.~Carpentier, J.~Sivic, and V.~Petrik, ``Physpose: Refining 6d object poses with physical constraints,'' \emph{arXiv preprint arXiv:2503.23587}, 2025.

\bibitem{macklin2022warp}
M.~Macklin, ``{Warp: A High-Performance Python Framework for GPU Simulation and Graphics},'' \url{https://github.com/NVIDIA/warp}, 2022, {NVIDIA}.

\bibitem{ravi2020pytorch3d}
N.~Ravi, J.~Reizenstein, D.~Novotny, T.~Gordon, W.-Y. Lo, J.~Johnson, and G.~Gkioxari, ``Accelerating 3d deep learning with pytorch3d,'' \emph{arXiv:2007.08501}, 2020.

\bibitem{isaac_sim1}
\BIBentryALTinterwordspacing
{NVIDIA}, ``{NVIDIA Isaac Sim},'' 2025. [Online]. Available: \url{https://github.com/isaac-sim/IsaacSim}
\BIBentrySTDinterwordspacing

\bibitem{7251504}
B.~Calli, A.~Singh, A.~Walsman, S.~Srinivasa, P.~Abbeel, and A.~M. Dollar, ``The ycb object and model set: Towards common benchmarks for manipulation research,'' in \emph{International Conference on Advanced Robotics (ICAR)}, 2015.

\bibitem{Panozzo2014LIBIGLAC}
A.~Jacobson, D.~Panozzo \emph{et~al.}, ``{libigl}: A simple {C++} geometry processing library,'' 2018, https://libigl.github.io/.

\bibitem{jin20256dope}
Y.~Jin, V.~Prasad, S.~Jauhri, M.~Franzius, and G.~Chalvatzaki, ``6dope-gs: Online 6d object pose estimation using gaussian splatting,'' in \emph{Proceedings of the IEEE/CVF International Conference on Computer Vision}, 2025, pp. 8032--8043.

\bibitem{liu2025gscpr}
\BIBentryALTinterwordspacing
C.~Liu, S.~Chen, Y.~S. Bhalgat, S.~HU, M.~Cheng, Z.~Wang, V.~A. Prisacariu, and T.~Braud, ``{GS}-{CPR}: Efficient camera pose refinement via 3d gaussian splatting,'' in \emph{The Thirteenth International Conference on Learning Representations}, 2025. [Online]. Available: \url{https://openreview.net/forum?id=mP7uV59iJM}
\BIBentrySTDinterwordspacing

\end{thebibliography}

\end{document}